\documentclass{article}

\usepackage[preprint]{neurips_2026}

\usepackage[utf8]{inputenc}
\usepackage[T1]{fontenc}
\usepackage{hyperref}
\hypersetup{hidelinks}
\usepackage{url}
\usepackage{booktabs}
\usepackage{amsmath,amssymb,amsfonts,amsthm}
\usepackage{mathtools}
\usepackage{nicefrac}
\usepackage{microtype}
\usepackage{xcolor}
\usepackage{graphicx}
\usepackage{array}
\usepackage{tabularx}
\usepackage{enumitem}
\setcitestyle{authoryear,round}

\newtheorem{definition}{Definition}
\newtheorem{assumption}{Assumption}
\newtheorem{theorem}{Theorem}
\newtheorem{proposition}{Proposition}

\newcommand{\E}{\mathbb{E}}
\newcommand{\R}{\mathbb{R}}

\newcommand{\norm}[1]{\left\lVert #1 \right\rVert}

\title{Algometrics: Forecasting Under Algorithmic Feedback}

\author{
  Marc Schmitt\thanks{Correspondence: marc.schmitt [at] cs.ox.ac.uk} \\
  University of Oxford \\
}

\begin{document}

\maketitle

\begin{abstract}
In algorithmic markets, predictive models become part of the data-generating process they aim to forecast. Once their outputs are converted into trades, allocations, execution schedules, or risk controls, they change the future data on which they are evaluated. I introduce algometrics, a framework for time series whose evolution depends on the predictive algorithms forecasting them. The framework distinguishes historical risk, measured under passive forecasting, from deployment risk, measured when forecasts drive actions. I prove three results. First, deployment risk is not identifiable from passive historical data alone: even in a one-step linear feedback model, infinitely many algorithm-mediated environments induce the same historical law while implying different deployment risks for the same forecaster. Second, historical model rankings can invert under crowding, so a predictor with lower passive error can have higher deployment error once similar algorithms are adopted. Third, randomized or instrumented actions identify short-horizon linear feedback, and I derive a finite-sample bound for deployment-risk estimation. These results suggest that time-series benchmarks in algorithmic markets should report feedback sensitivity alongside predictive accuracy.
\end{abstract}

\section{Introduction}

Financial markets are a natural destination for machine learning because they are high-dimensional, weakly predictive, and noisy. A growing empirical and theoretical literature argues that model complexity can be valuable in return prediction: many weak signals, interactions, and nonlinearities may be invisible to parsimonious models but useful to modern ML systems \citep{gu2020empirical,kelly2024virtue}. Finance is not just a domain where overfitting is easy; it is one where controlled complexity can be productive.

This paper studies a different source of complexity. In financial markets, forecasts are rarely inert. A return prediction becomes an order or a portfolio weight; a credit score changes funding access; a liquidity forecast changes execution; a risk model changes leverage and margin; a widely copied signal changes the price process on which future models train. The statistical object is a time series partly generated by the algorithms that forecast it.

I call this problem \emph{algometrics}: the measurement and learning of time series under algorithmic feedback. It refers to settings in which (i) predictive algorithms observe a history, (ii) their outputs are converted into actions, (iii) actions influence future observations, and (iv) a population of related algorithms may create crowding or strategic adaptation. Financial markets are the central example; the same pattern arises in online advertising, pricing, recommender systems, platform labor, and cybersecurity.

The paper makes three claims. First, historical risk and deployment risk are different estimands. A model can be accurate when it passively forecasts a market and inaccurate when it becomes part of the market's order flow. Second, the difference is not a small-sample nuisance. Without variation in algorithmic actions, deployment risk is generally not identifiable from passive histories. Third, evaluation is still possible if the data include interventions, randomized exposure, simulator access with calibrated impact, or other instruments that reveal how outcomes respond to algorithmic actions.

The contribution is an identification argument. I formalize an \emph{algorithm-mediated time series} as a sequence whose transition kernel depends on algorithm-induced actions. I define the \emph{feedback gap} between passive and deployment risk. I then prove two negative results and one positive result. The first negative result shows non-identifiability: in a one-step linear feedback model, all values of the feedback coefficient induce the same passive historical distribution, but they imply different deployment risks. The second shows ranking inversion: the historically best predictor can become worse than a conservative predictor after crowding. The positive result shows that randomized actions identify a short-horizon linear feedback coefficient and yield a finite-sample deployment-risk bound.

The contribution type is theory. The closed-form illustration in Section~\ref{sec:simulation} visualizes the mechanism; the main claims are mathematical. The message is that time-series benchmarks in algorithmic environments should report historical prediction scores together with the assumptions under which those scores extrapolate to deployment.

\section{From statistical complexity to algorithmic feedback}
\label{sec:motivation}

The case for complex ML in finance is often framed as a response to parsimony. Classical equity-premium prediction struggled to beat simple baselines \citep{welch2008comprehensive}. Recent ML work has revisited that conclusion by exploiting large cross-sections, nonlinearities, interactions, and regularization \citep{gu2020empirical}. \citet{kelly2024virtue} argue more directly for a ``virtue of complexity'' in return prediction: complex models can recover predictive structure that small models miss. Skeptical perspectives note that ML performance in finance is sensitive to economic restrictions \citep{avramov2023ml,martin2022market} and to multiple-testing concerns \citep{harvey2016luck}, and tree-based approaches deliver gains comparable to deep models with greater interpretability \citep{bryzgalova2025forest}. Algometrics is complementary to both sides of this debate: even if the predictive case for complexity is granted, the deployment-relevant case additionally requires modeling how the predictor changes the prediction target.

Algometrics adds an endogenous layer. A model's complexity affects not only what it learns from history but also how it acts on the market. A high-capacity model may trade more aggressively, coordinate unintentionally with models trained on similar data, or produce signals whose value decays once others imitate them. These effects are not captured by the usual historical loss
\begin{equation}
  R_0(f)=\E\big[\ell(f(H_t),Y_{t+1})\big],
\end{equation}
where $H_t$ is the observed history and $Y_{t+1}$ is the next return, price change, spread, default indicator, or other target. The loss treats the target as external to the forecast.

The concern is related to performative prediction, where predictions influence the distribution of outcomes \citep{perdomo2020performative, mendler2020stochastic, miller2021outside}. It also connects to strategic classification \citep{hardt2016strategic}, market impact \citep{kyle1985continuous,almgren2001optimal}, adaptive markets \citep{lo2004adaptive}, herding \citep{cont2000herd}, and agent-based computational finance \citep{lebaron2006agent}. The distinct feature here is the time-series estimand: historical data are often collected under one algorithmic regime, while deployment occurs under another.

\begin{table}[t]
\caption{Core objects in algometrics. The terms are operational: each points to an estimand or diagnostic for time-series learning under feedback.}
\label{tab:objects}
\centering
\small
\begin{tabularx}{\linewidth}{>{\raggedright\arraybackslash}p{0.29\linewidth}>{\raggedright\arraybackslash}p{0.64\linewidth}}
\toprule
Object & Meaning \\
\midrule
Historical risk $R_0(f)$ & Prediction loss when the model forecasts a sequence without changing the transition law. \\
Deployment risk $R_m(f)$ & Prediction loss when the model is deployed with adoption or capital share $m$ and its outputs affect future observations. \\
Feedback gap $\Gamma_m(f)$ & Difference $R_m(f)-R_0(f)$, or the analogous gap for a reward score. \\
Algorithmic elasticity & Sensitivity of the next-observation distribution to algorithm-induced actions. \\
Crowding curve & Deployment performance as a function of the mass of agents using correlated predictors. \\
Instrumented evaluation & Evaluation using randomized, staggered, or otherwise exogenous variation in algorithmic actions. \\
\bottomrule
\end{tabularx}
\end{table}

This framing does not deny the value of historical benchmarks. They are essential for detecting leakage, survivorship bias, overfitting, and unrealistic cost assumptions. The claim is narrower: a benchmark that estimates $R_0(f)$ should not be interpreted as estimating $R_m(f)$ unless the feedback gap is argued to be negligible, bounded, or measured.

\section{Algorithm-mediated time series}
\label{sec:framework}

Let $H_t=(Y_1,A_1,\ldots,Y_{t-1},A_{t-1},Y_t)$ denote the observed history up to and including the time-$t$ outcome but before the time-$t$ action. A forecaster $f\in\mathcal{F}$ maps histories to predictions $\hat Y_{t+1}=f(H_t)$. A deployment map $\pi$ converts a forecast into an action $A_t=\pi(f,H_t,m_t)$, where $m_t\in[0,1]$ is the adoption, capital, or traffic share affected by the forecaster. The convention orders observation, prediction, and action within each period: $Y_t$ is observed first, $f$ produces $\hat Y_{t+1}$ from $H_t$, and $A_t$ is then induced; $A_t$ enters $H_{t+1}$, not $H_t$. In a market, $A_t$ may represent demand, order size, portfolio weight, leverage, cancellation intensity, or an execution schedule.

\begin{definition}[Algorithm-mediated time series]
An algorithm-mediated time series is a family of transition kernels
\begin{equation}
  P_\eta\big(Y_{t+1}\in dy, S_{t+1}\in ds \mid H_t,S_t,A_t,m_t\big),
\end{equation}
where $S_t$ is a latent or observed state, $A_t$ is an action induced by one or more algorithms, $m_t$ is an adoption mass, and $\eta$ indexes environment parameters such as market impact, liquidity, or strategic response. The passive regime sets $A_t=0$ and $m_t=0$. A deployment regime uses $A_t=\pi(f,H_t,m_t)$.
\end{definition}

For a loss $\ell$, historical risk is
\begin{equation}
  R_0(f)=\E_{P_\eta^{0}}\left[\ell(f(H_t),Y_{t+1})\right],
\end{equation}
where $P_\eta^0$ is the law induced by passive observation. Deployment risk at adoption $m$ is
\begin{equation}
  R_m(f)=\E_{P_\eta^{f,m}}\left[\ell(f(H_t),Y_{t+1})\right],
\end{equation}
where $P_\eta^{f,m}$ is the law induced when $f$ is acted upon. The feedback gap is
\begin{equation}
  \Gamma_m(f)=R_m(f)-R_0(f).
\end{equation}
For reward objectives, such as expected return or Sharpe-like scores, one can define the gap with the sign reversed. The mathematical issue is the same: a passive score and a deployment score are different estimands. The trajectory definition above admits a useful one-step specialization: the \emph{one-step deployment risk} is
\begin{equation}
\label{eq:onestep}
  R_m^{(1)}(f)=\E_{H_t\sim P_\eta^0}\!\left[\E\!\left[\ell(f(H_t),Y_{t+1})\,\big|\,H_t,\,A_t=\pi(f,H_t,m)\right]\right],
\end{equation}
where the outer expectation draws histories from the passive law and only the time-$t$ action is induced by the deployment policy. Theorems~\ref{thm:nonid} and~\ref{thm:ranking}, and Proposition~\ref{prop:smallfeedback}, are statements about $R_m^{(1)}(f)$. 
Comparing deployment risks across forecasters is a policy comparison: each forecaster $f$ is evaluated under the data-generating process $P_\eta^{f,m}$ induced by its own deployment, not on a common realized label sequence. This is not a confound; it is the object of study. In algorithmic environments the future depends on the policy, so a deployment-risk comparison is necessarily a comparison of policies and of the worlds they induce.

A useful local diagnostic is algorithmic elasticity. For a metric $d$ on distributions,
\begin{equation}
  \mathcal{E}(H_t;a,a')=\frac{d\left(P_\eta(Y_{t+1}\mid H_t,A_t=a),\,P_\eta(Y_{t+1}\mid H_t,A_t=a')\right)}{\norm{a-a'}},
\end{equation}
whenever $a\neq a'$, where $P_\eta(Y_{t+1}\mid H_t,A_t)=\int P_\eta(Y_{t+1}\mid H_t,S_t,A_t)\,P_\eta(dS_t\mid H_t)$ marginalizes the latent state under the prevailing regime. High elasticity means that the target is sensitive to actions induced by forecasts. In finance, this can be caused by price impact, liquidity withdrawal, signal crowding, or strategic response. In low-elasticity regimes, historical risk may be a reasonable proxy for deployment risk. In high-elasticity regimes, the proxy is suspect.

\subsection{When passive evaluation is justified}
\label{sec:small-feedback}

The framework does not imply that every historical benchmark is invalid. It identifies the condition under which historical evaluation is doing more than passive prediction: the action-induced distributional change must be small relative to the loss. A simple bound makes this explicit.

\begin{proposition}[Small-feedback bound]
\label{prop:smallfeedback}
Fix a one-step evaluation conditional on histories drawn from the passive law. Fix a ground metric $d_Y$ on the outcome space and let $W_1$ denote the corresponding $1$-Wasserstein distance. Suppose the loss $\ell(\hat y,y)$ is $L_\ell$-Lipschitz in $y$ with respect to $d_Y$, and that
\begin{equation}
  W_1\left(P(Y_{t+1}\mid H_t,A_t=a),P(Y_{t+1}\mid H_t,A_t=0)\right)
  \leq \kappa(H_t)\norm{a}
\end{equation}
for all feasible actions $a$. Writing $\pi_f(H_t):=\pi(f,H_t,m)$ for the deployment policy at fixed adoption $m$ and $\Gamma(f):=\Gamma_m^{(1)}(f)$ for the one-step feedback gap, we have
\begin{equation}
  |\Gamma(f)| \leq L_\ell\,\E\left[\kappa(H_t)\norm{\pi_f(H_t)}\right].
\end{equation}
\end{proposition}

The proposition is useful because it separates two questions that are often mixed together. A model can be statistically complex while having small market footprint, in which case passive evaluation may be adequate. Conversely, a simple model can have a large feedback gap if it controls enough capital or traffic. Complexity of representation and complexity of deployment are distinct axes.

\section{Passive histories do not identify deployment risk}
\label{sec:nonidentification}

The first result shows that the feedback gap cannot in general be recovered from passive histories. The theorem is intentionally elementary. It is stronger to show failure in a one-step linear model than to rely on complicated market dynamics.

\begin{theorem}[Passive non-identifiability]
\label{thm:nonid}
Consider the one-step linear feedback family
\begin{equation}
  Y_{t+1}=\mu(H_t)+\beta A_t+\varepsilon_{t+1}, \qquad \E[\varepsilon_{t+1}\mid H_t,A_t]=0,
\end{equation}
with squared loss and finite second moments. In the passive regime $A_t=0$. Fix any forecaster $f$ and deployment policy $A_t=\pi_f(H_t)$ satisfying $\E[\pi_f(H_t)^2]>0$ under the passive history law. Then all values of $\beta$ induce the same passive distribution over $(H_t,Y_{t+1})$, but the one-step deployment risk of $f$ is a nonconstant quadratic function of $\beta$. Consequently, no estimator based only on passive histories can identify deployment risk uniformly over this family.
\end{theorem}

\noindent The proof is in Appendix~\ref{app:proofs}. The intuition is that passive data only reveal $\mu(H_t)$ because $A_t$ never varies. The coefficient $\beta$ is invisible until an action is taken. The result is structurally the time-series and feedback analog of \citet{manski1993identification}'s reflection problem: observed outcomes alone cannot separate the model's effect on the world from the world's effect on the model's inputs without exogenous variation.

\begin{figure}[t]
\centering
\includegraphics[width=\linewidth]{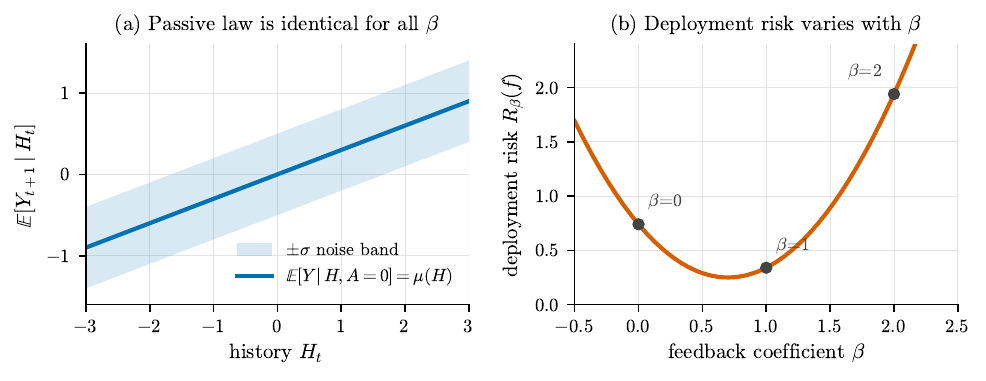}
\caption{Closed-form illustration of Theorem~\ref{thm:nonid}. In panel (a), passive observations are generated with $A=0$, so the conditional law $Y\mid H, A=0$ is the same for all feedback coefficients $\beta$. Passive data therefore reveal $\mu(H)$ but not the action effect. In panel (b), the same forecaster $f(h)=h$ is deployed with action $A=f(H)$. The deployment risk $R_\beta(f)$ varies with $\beta$, even though the passive law in panel (a) is unchanged. Thus passive historical data do not identify deployment risk.}
\label{fig:nonid}
\end{figure}

Figure~\ref{fig:nonid} visualizes this concretely: the same observable passive law is consistent with infinitely many feedback coefficients, each producing a different deployment risk for the same forecaster.

The theorem does not claim that feedback is large in every financial model. It says that passive data cannot identify the feedback coefficient: if the evaluated action was absent from the history, or present at a different adoption share, additional passive observation will not recover it. The estimand on offer is the wrong one.

\section{Historical rankings can invert after crowding}
\label{sec:ranking}

The second result shows that feedback can change model rankings. This matters because benchmarks usually select models by comparing scores. If a leaderboard ranks passive risk but users care about deployment risk, the selected model can be the wrong one.

\begin{theorem}[Crowding-induced ranking inversion]
\label{thm:ranking}
Let $H\sim N(0,1)$ and $\varepsilon\sim N(0,\sigma^2)$ be independent. Under passive observation,
\begin{equation}
  Y^0=H+\varepsilon.
\end{equation}
For $c\in[0,1]$, define the forecaster $f_c(H)=cH$ and suppose deployment induces action $A=\alpha f_c(H)$ with adoption intensity $\alpha\geq 0$. Let the deployed target be
\begin{equation}
  Y^{c,\alpha}=H-\gamma \alpha cH+\varepsilon,
\end{equation}
where $\gamma>0$ is a crowding or negative-impact coefficient. Then for every $c\in[0,1]$,
\begin{equation}
\label{eq:ranking-closed-form}
  R_0(f_c)=(c-1)^2+\sigma^2, \qquad R_\alpha(f_c)=\big(c(1+\gamma\alpha)-1\big)^2+\sigma^2.
\end{equation}
Fix any conservative predictor $f_{c'}$ with $c'\in[0,1)$. Then $f_1$ has strictly lower passive risk than $f_{c'}$,
\begin{equation}
  R_0(f_1)=\sigma^2 < (c'-1)^2+\sigma^2=R_0(f_{c'}),
\end{equation}
but $f_1$ has strictly higher deployment risk than $f_{c'}$ whenever
\begin{equation}
\label{eq:ranking-threshold}
  \gamma\alpha > \frac{1-c'}{1+c'}.
\end{equation}
The case $c'=0$ recovers the threshold $\gamma\alpha>1$. The case $c'=0.25$ used in Figure~\ref{fig:gap} gives the threshold $\gamma\alpha>0.6$.
\end{theorem}

This stylized theorem captures a documented market phenomenon. Published anomalies systematically lose predictive power once they become widely known \citep{mclean2016academic,falck2022when}, and the August 2007 quant crisis is the canonical real-world example of correlated strategies failing simultaneously when many funds deploy similar signals \citep{khandani2011quants}. The accurate signal is valuable when it is passive and scarce; if enough capital trades on it, the induced demand can move prices against the signal, consume liquidity, or accelerate the decay of the opportunity. A conservative predictor that looks worse in passive loss can be safer under adoption. The theorem is not about the particular linear form. It shows why a single historical score is an incomplete decision rule for model selection.

\section{A positive result: instrumented feedback estimation}
\label{sec:positive}

The negative results do not imply that deployment risk is unknowable. They imply that some variation in actions is needed. Such variation can come from randomized order slicing, staggered deployment, A/B exposure, exogenous instrumented demand, or calibrated simulators. The result below is in spirit an instrumented-identification statement \citep{imbens1994late, harris2022strategic}, in which exogenous variation in actions identifies a causal feedback coefficient that passive observation cannot recover. A complementary line of work studies online regret minimization under performative feedback \citep{jagadeesan2022regret,izzo2021how}, where the learner controls the deployment sequence directly. I give a finite-sample version for a short-horizon linear feedback model.

\begin{assumption}[Instrumented linear feedback]
\label{ass:linear}
For $i=1,\ldots,n$, observe $Y_i=z_i^\top w+\varepsilon_i$ where $z_i=(\phi_i,A_i)\in\R^p$, $w=(\theta,\beta)\in\R^p$, and $\norm{z_i}_2\leq L$ almost surely. Let $\{\mathcal{F}_i\}_{i\geq 0}$ be a filtration such that $z_i$ is $\mathcal{F}_{i-1}$-measurable and the action $A_i$ is randomized or instrumented conditional on $\mathcal{F}_{i-1}$. The errors form a martingale difference sequence with $\E[\varepsilon_i\mid\mathcal{F}_{i-1}]=0$, $\mathrm{Var}(\varepsilon_i\mid\mathcal{F}_{i-1})=\sigma^2$, and $\varepsilon_i\mid\mathcal{F}_{i-1}$ is $\sigma^2$-sub-Gaussian. The analysis works conditionally on the design event $\mathcal{G}_\lambda=\{G_n\succeq \lambda I_p\}$, where $G_n=n^{-1}\sum_i z_i z_i^\top$ and $\lambda>0$. This MDS formulation is appropriate for autoregressive time-series settings in which $z_i$ contains lags of $Y$ and conditioning on the full design $\{z_j\}_{j=1}^n$ would entangle future covariates with past noise.
\end{assumption}

\begin{theorem}[Finite-sample deployment-risk estimation]
\label{thm:positive}
Under Assumption~\ref{ass:linear}, let $\hat w$ be ordinary least squares. Conditional on the design event $\mathcal{G}_\lambda$, with probability at least $1-\delta$ over the regression noise,
\begin{equation}
  \norm{\hat w-w}_2\leq \frac{\sigma L}{\lambda}\sqrt{\frac{2p\log(2p/\delta)}{n}}.
\end{equation}
For any deployment policy with feature-action vector $z^\pi(H)$ satisfying $\norm{z^\pi(H)}_2\leq L$ almost surely, the induced conditional mean $z^\pi(H)^\top w$ is estimated uniformly up to
\begin{equation}
  \left| z^\pi(H)^\top(\hat w-w)\right|\leq \epsilon_n,
  \qquad \epsilon_n\;:=\;\frac{\sigma L^2}{\lambda}\sqrt{\frac{2p\log(2p/\delta)}{n}}.
\end{equation}
Suppose further that the oracle prediction error $|f(H)-z^\pi(H)^\top w|$ is bounded by $B$ almost surely. Then the plug-in squared-loss deployment-risk estimate
\begin{equation*}
  \widehat R_m(f)\;=\;\E_H\!\left[(f(H)-z^\pi(H)^\top\hat w)^2\right]+\sigma^2
\end{equation*}
differs from the oracle deployment risk by at most
\begin{equation}
  \left|\widehat R_m(f)-R_m(f)\right|\;\leq\;2B\epsilon_n+\epsilon_n^2.
\end{equation}
\end{theorem}

The theorem is modest by design. It does not solve market simulation or long-horizon equilibrium. It says that once actions vary exogenously enough to identify their effect, deployment-sensitive risk estimation becomes an ordinary statistical problem, of the kind that double/debiased ML methods address routinely \citep{chernozhukov2018dml}. Recent work on causal estimation of performativity in cross-sectional and non-autoregressive settings \citep{mendler2022anticipating, cheng2024causal} pursues a related identifiability question. The contrast with Theorem~\ref{thm:nonid} is the main point: passive histories hide feedback; instrumented histories can reveal it.

\paragraph{Misspecified feedback.}
The linear specification in Assumption~\ref{ass:linear} is a baseline. Empirical microstructure exhibits concave impact, with the square-root law $\Delta p\propto\mathrm{sign}(Q)\sqrt{|Q|}$ as a robust regularity \citep{almgren2005direct}. If the true conditional mean is $g(z)$ with bounded misspecification $\rho:=\sup_z|g(z)-z^\top w_\star|$ relative to the best linear projection $w_\star$, the parameter bound in Theorem~\ref{thm:positive} carries an additive bias of order $\rho/\lambda$, and the plug-in deployment-risk estimate inflates by $O(B\rho+\rho^2)$. Sieve regression, kernel methods, or parametric concave-impact models are natural extensions that recover identification under richer feedback while preserving the basic mechanism.

\section{Illustrative closed-form example}
\label{sec:simulation}

\begin{figure}[t]
\centering
\includegraphics[width=0.75\linewidth]{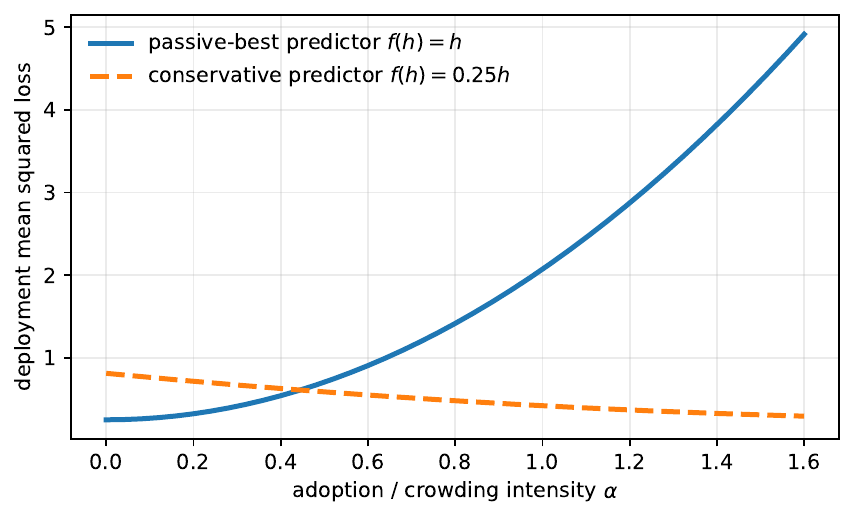}
\caption{
Closed-form illustration of Theorem~\ref{thm:ranking} with \(\sigma=0.5\) and \(\gamma=1.35\), comparing the passive-best predictor \(f(h)=h\) with the conservative predictor \(f(h)=0.25h\). Each curve reports deployment risk under the data-generating process induced by that predictor's own action; the comparison is therefore a comparison of policies, not forecasts on a common label sequence. The passive-best predictor has the lowest historical risk at \(\alpha=0\), but its deployment risk rises with adoption, producing a ranking inversion at \(\alpha^\star\approx 0.44\).
}
\label{fig:gap}
\end{figure}

Figure~\ref{fig:gap} visualizes the ranking inversion in Theorem~\ref{thm:ranking} with conservative parameter $c'=0.25$ and feedback coefficient $\gamma=1.35$, so the inversion threshold $\gamma\alpha>(1-c')/(1+c')=0.6$ is crossed at $\alpha\approx 0.44$. Let $H\sim N(0,1)$ and $\varepsilon\sim N(0,0.5^2)$. The passive target is $Y^0=H+\varepsilon$. The passive-best model predicts $f(H)=H$. A conservative model predicts $0.25H$. At deployment, adoption intensity $\alpha$ creates negative feedback $Y^{c,\alpha}=H-1.35\alpha cH+\varepsilon$. The passive-best predictor remains superior for small $\alpha$, but its loss rises rapidly as adoption increases because its own action changes the target. The conservative predictor has worse passive fit but lower feedback exposure.

The example uses no proprietary data and is not intended as a market model. It is a diagnostic example for benchmark design. A financial time-series benchmark that reports only the left endpoint of the figure is answering a passive question. A deployment-aware benchmark would report a crowding curve, a feedback assumption, or an instrumented estimate of action sensitivity.

\section{Implications for financial ML evaluation}
\label{sec:implications}

The theory suggests several evaluation practices. Financial ML papers should specify whether their claims concern passive prediction, paper trading, small-scale deployment, or population-level deployment, since these are different regimes. Benchmarks should report adoption-sensitivity curves when a model's actions plausibly affect the target, and historical leaderboard rankings should be stress-tested under impact, crowding, and adaptive-opponent scenarios. Existing tooling can support this: agent-based market simulators such as ABIDES and ABIDES-Gym make interaction explicit \citep{byrd2019abides,amrouni2021abidesgym}, and platforms such as Qlib and FinRL can be extended with feedback modules rather than treated as historical replay systems alone \citep{yang2020qlib,liu2020finrl,liu2022finrlmeta}. Finally, when real deployments are possible, randomized or staggered exposure should be valued as measurement infrastructure, not as product experimentation alone.

Table~\ref{tab:diagnostics} translates the theory into benchmark diagnostics. None of these diagnostics requires a perfect market simulator. The goal is to make the assumed deployment regime explicit and to test whether the main model ranking survives plausible feedback perturbations.

\begin{table}[t]
\caption{Deployment-aware diagnostics suggested by the theory.}
\label{tab:diagnostics}
\centering
\small
\begin{tabularx}{\linewidth}{>{\raggedright\arraybackslash}p{0.28\linewidth}>{\raggedright\arraybackslash}p{0.34\linewidth}>{\raggedright\arraybackslash}p{0.28\linewidth}}
\toprule
Diagnostic & Measurement idea & Failure mode revealed \\
\midrule
Crowding curve & Recompute performance over adoption or capital share $m$. & A signal works only when few agents use it. \\
Impact perturbation & Add calibrated price-impact or liquidity-cost response to actions. & High-turnover predictors look good only under historical replay. \\
Action randomization & Randomize small exposure, routing, or timing where ethically and legally feasible. & Deployment risk diverges from passive estimates due to unmeasured feedback. \\
Adaptive opponent & Evaluate against agents that learn from induced order flow or public signals. & The model is brittle to strategic response. \\
Population stress & Simulate correlated policies rather than isolated models. & Individually safe models create collective fragility. \\
\bottomrule
\end{tabularx}
\end{table}

These suggestions are not limited to trading. In credit, a model can change default labels by changing access to credit. In insurance, pricing models can change the risk pool. In recommender systems, ranking models change attention and future engagement labels. In cybersecurity, detection models change attacker behavior. Finance is simply the cleanest laboratory because actions, prices, and feedback are often recorded at fine temporal resolution.

\section{Related work}
\label{sec:related}

\paragraph{Financial ML and return prediction.}
Modern empirical asset pricing has shown that flexible models can improve the prediction of returns and characteristics-based portfolios \citep{gu2020empirical}. The ``virtue of complexity'' argument gives a theoretical and empirical rationale for using many parameters in return prediction \citep{kelly2024virtue}. Algometrics is complementary: it asks when a learned signal remains valid after the model changes the market state through action.

\paragraph{Performative and strategic prediction.}
Performative prediction formalizes the idea that predictions can change the distribution on which they are evaluated \citep{perdomo2020performative,mendler2020stochastic,miller2021outside}, with extensions to stateful environments \citep{brown2022stateful}, multi-agent decision-dependent games \citep{narang2023multiplayer}, the influence exerted by deployed predictors on the data they observe \citep{hardt2022performative,mendlerdunner2024measuring}, and the broader agenda surveyed in \citet{hardt2023performative}. Strategic classification studies agents who adapt their features in response to a classifier \citep{hardt2016strategic,levanon2021strategic}. A recent application to financial market making is developed in \citet{kleitsikas2025performative}. Relative to this literature, algometrics emphasizes a time-series identification problem: the adapted object is the sequence of future labels, not an individual feature vector, and historical data are typically collected under a different algorithmic regime than deployment.

\paragraph{Market microstructure and agent-based finance.}
Market impact, liquidity, and execution costs are central to microstructure \citep{kyle1985continuous,almgren2001optimal}. Agent-based computational finance studies aggregate outcomes generated by interacting traders \citep{lebaron2006agent,cont2000herd}. High-fidelity simulators such as ABIDES make these interactions programmable for ML research \citep{byrd2019abides,amrouni2021abidesgym}. The contribution is not a new simulator; it is an identification argument for why passive time-series benchmarks cannot, by themselves, estimate deployment risk.

\section{Limitations and scope}
\label{sec:limitations}

The framework is intentionally stylized. The theorems use one-step or short-horizon feedback rather than full equilibrium markets. Real markets include latent information, heterogeneous objectives, inventory constraints, asymmetric information, regulatory limits, and adversarial behavior. The positive result assumes randomized or instrumented actions and a linear feedback model; many deployments violate both assumptions. The closed-form illustration is a visualization of the mechanism, not evidence about any real asset class. 

The term algometrics should also not be read as a replacement for econometrics, market microstructure, or time-series analysis. It names a subset of problems in which algorithms are part of the data-generating process. Classical tools remain necessary. The new requirement is to state when passive historical risk is a deployment-relevant estimand and when it is only a first-stage diagnostic. Finally, deployment-sensitive evaluation can be misused. Better estimates of feedback may improve risk management and benchmark validity, but they may also help institutions trade more effectively against less sophisticated participants. The response is to make assumptions and externalities visible, especially when models are released as reusable financial ML infrastructure.

\section{Conclusion}

The virtue of complexity in financial ML concerns what rich models can learn from high-dimensional markets. Algometrics asks what happens after those models act. The paper formalized this distinction through algorithm-mediated time series, historical risk, deployment risk, and the feedback gap. It showed that passive histories do not identify deployment risk in general, that model rankings can invert under crowding, and that instrumented action variation can recover short-horizon feedback in a linear setting. In algorithmic markets, time-series learning should measure how the future responds to the models that predict it.

{\small
\bibliographystyle{plainnat}
\bibliography{bib}
}

%%%%%%%%%%%%%%%%%%%%%%%%%%%%%%%%%%%%%%%%%%%%%%%%%%%%%%%%%%%%

\appendix

\section{Proofs}
\label{app:proofs}

\subsection{Proof of Proposition~\ref{prop:smallfeedback}}

For a fixed history $H_t$ and prediction $\hat y=f(H_t)$, the map $y\mapsto \ell(\hat y,y)$ is $L_\ell$-Lipschitz. By the Kantorovich-Rubinstein dual characterization of $W_1$,
\begin{align}
 &\left|\E[\ell(\hat y,Y)\mid H_t,A_t=a]
       -\E[\ell(\hat y,Y)\mid H_t,A_t=0]\right| \\
 &\hspace{0.4in}\leq L_\ell W_1\left(P(Y\mid H_t,A_t=a),
                       P(Y\mid H_t,A_t=0)\right).
\end{align}
Using the assumed elasticity bound with $a=\pi_f(H_t)$ and taking expectations over passive histories gives
\begin{equation}
  |\Gamma(f)|\leq L_\ell\E[\kappa(H_t)\norm{\pi_f(H_t)}].
\end{equation}
This proves the claim. \qed

\subsection{Proof of Theorem~\ref{thm:nonid}}

In the passive regime $A_t=0$, the model is
\begin{equation}
  Y_{t+1}=\mu(H_t)+\varepsilon_{t+1}.
\end{equation}
The passive conditional law of $Y_{t+1}$ given $H_t$ is therefore independent of $\beta$. Hence any two values $\beta$ and $\beta'$ induce the same passive distribution over $(H_t,Y_{t+1})$.

Under deployment, $A_t=\pi_f(H_t)$ and
\begin{equation}
  Y_{t+1}^{\beta}=\mu(H_t)+\beta \pi_f(H_t)+\varepsilon_{t+1}.
\end{equation}
The one-step deployment risk of $f$ under squared loss is
\begin{align}
  R_\beta(f)
  &=\E\left[\left(f(H_t)-\mu(H_t)-\beta\pi_f(H_t)-\varepsilon_{t+1}\right)^2\right] \\
  &=\E\left[\left(f(H_t)-\mu(H_t)-\varepsilon_{t+1}\right)^2\right] \\
  &\quad -2\beta\E\left[\pi_f(H_t)\left(f(H_t)-\mu(H_t)\right)\right]
    +\beta^2\E\left[\pi_f(H_t)^2\right],
\end{align}
where the cross-term involving $\varepsilon_{t+1}$ vanishes by the tower property since $\E[\varepsilon_{t+1}\mid H_t,A_t]=0$.
The coefficient on $\beta^2$ is positive by assumption, so $R_\beta(f)$ is a nonconstant quadratic function of $\beta$. Since the passive data distribution is identical for all $\beta$ but the deployment risk differs for at least two values of $\beta$, no estimator that is a measurable function only of passive histories can identify deployment risk uniformly over this family. \qed

\subsection{Proof of Theorem~\ref{thm:ranking}}

For $f_c(H)=cH$ under passive observation,
\begin{equation}
  R_0(f_c)=\E\left[(cH-H-\varepsilon)^2\right]=(c-1)^2+\sigma^2.
\end{equation}
Under deployment of $f_c$, the target is $Y^{c,\alpha}=H-\gamma\alpha cH+\varepsilon$, and the forecast is $cH$, so
\begin{align}
  R_\alpha(f_c)
  &=\E\left[\left(cH-(H-\gamma\alpha cH+\varepsilon)\right)^2\right] \\
  &=\E\left[\left((c(1+\gamma\alpha)-1)H-\varepsilon\right)^2\right] \\
  &=(c(1+\gamma\alpha)-1)^2+\sigma^2,
\end{align}
which establishes equation~\eqref{eq:ranking-closed-form}.

Fix $c'\in[0,1)$. The passive comparison is immediate: $R_0(f_1)=\sigma^2<(c'-1)^2+\sigma^2=R_0(f_{c'})$. For deployment risk,
\begin{equation}
  R_\alpha(f_1)>R_\alpha(f_{c'})
  \;\iff\;
  \gamma^2\alpha^2 > \big(c'(1+\gamma\alpha)-1\big)^2.
\end{equation}
It remains to resolve the absolute value on the right-hand side.

\emph{Regime I: $c'(1+\gamma\alpha)\leq 1$.} The right-hand side equals $(1-c'(1+\gamma\alpha))^2$, and taking nonnegative square roots gives
\begin{equation}
  \gamma\alpha > 1-c'-c'\gamma\alpha,
  \quad\text{i.e.,}\quad
  \gamma\alpha(1+c')>1-c',
\end{equation}
which is equivalent to~\eqref{eq:ranking-threshold}.

\emph{Regime II: $c'(1+\gamma\alpha)>1$.} Then $c'(1+\gamma\alpha)-1>0$, and the inequality $\gamma^2\alpha^2>(c'(1+\gamma\alpha)-1)^2$ is equivalent to $\gamma\alpha>c'(1+\gamma\alpha)-1$, i.e., $\gamma\alpha(1-c')>-(1-c')$, which is automatic for $c'\in[0,1)$ and $\gamma\alpha\geq 0$. Moreover, Regime II requires $\gamma\alpha>(1-c')/c'$, which strictly exceeds the threshold $(1-c')/(1+c')$ in~\eqref{eq:ranking-threshold}, so the inversion is already established before entering Regime II.

Setting $c'=0$ collapses Regime II to the empty set and recovers $\gamma\alpha>1$ from Regime I. \qed

\subsection{Proof of Theorem~\ref{thm:positive}}

Let $Z\in\R^{n\times p}$ have rows $z_i^\top$, let $Y=Zw+\varepsilon$, and let $\hat w=(Z^\top Z)^{-1}Z^\top Y$. Then
\begin{equation}
  \hat w-w=(Z^\top Z)^{-1}Z^\top\varepsilon.
\end{equation}
Since $G_n=n^{-1}Z^\top Z\succeq \lambda I_p$, we have $\norm{(Z^\top Z/n)^{-1}}_{\mathrm{op}}\leq 1/\lambda$. Therefore
\begin{equation}
  \norm{\hat w-w}_2\leq \frac{1}{\lambda}\left\|\frac{1}{n}Z^\top\varepsilon\right\|_2.
\end{equation}
For coordinate $j$, $(n^{-1}Z^\top\varepsilon)_j=n^{-1}\sum_i z_{ij}\varepsilon_i$ is a sum of martingale differences: $z_{ij}$ is $\mathcal{F}_{i-1}$-measurable with $|z_{ij}|\leq\norm{z_i}_2\leq L$, and $\varepsilon_i\mid\mathcal{F}_{i-1}$ is $\sigma^2$-sub-Gaussian, so $z_{ij}\varepsilon_i\mid\mathcal{F}_{i-1}$ is $z_{ij}^2\sigma^2$-sub-Gaussian. By the Azuma--Hoeffding inequality for sub-Gaussian martingale differences, $\sum_i z_{ij}\varepsilon_i$ is $n\sigma^2 L^2$-sub-Gaussian, so $(n^{-1}Z^\top\varepsilon)_j$ has proxy variance at most $\sigma^2L^2/n$. A union bound over $p$ coordinates gives, with probability at least $1-\delta$,
\begin{equation}
  \left\|\frac{1}{n}Z^\top\varepsilon\right\|_\infty
  \leq \sigma L\sqrt{\frac{2\log(2p/\delta)}{n}}.
\end{equation}
Since $\norm{v}_2\leq\sqrt{p}\norm{v}_\infty$, the stated parameter bound follows. For any policy vector $z^\pi(H)$ with norm at most $L$,
\begin{equation}
  |z^\pi(H)^\top(\hat w-w)|\leq L\norm{\hat w-w}_2=\epsilon_n.
\end{equation}
For the plug-in risk bound, fix $H$ and write $\mu_\pi=z^\pi(H)^\top w$, $\hat\mu_\pi=z^\pi(H)^\top\hat w$. Expanding,
\begin{equation}
  (f(H)-\hat\mu_\pi)^2-(f(H)-\mu_\pi)^2
  \;=\;2(f(H)-\mu_\pi)(\mu_\pi-\hat\mu_\pi)+(\mu_\pi-\hat\mu_\pi)^2.
\end{equation}
Using $|f(H)-\mu_\pi|\leq B$ almost surely and $|\mu_\pi-\hat\mu_\pi|\leq \epsilon_n$ from the previous display,
\begin{equation}
  \left|(f(H)-\hat\mu_\pi)^2-(f(H)-\mu_\pi)^2\right|\;\leq\;2B\epsilon_n+\epsilon_n^2.
\end{equation}
Taking expectations over $H$ and noting that the noise variance $\sigma^2$ cancels between $\widehat R_m(f)$ and $R_m(f)$ yields the stated bound. \qed

\end{document}